\documentclass[runningheads]{llncs}
\usepackage{graphicx}
\usepackage{multirow}
\usepackage{todonotes}
\usepackage{bm}
\usepackage{textcomp}
\usepackage{siunitx}
\usepackage{amsfonts}
\usepackage{amsmath}
\begin{document}
\title{Neural Multi-Scale Self-Supervised Registration for Echocardiogram Dense Tracking} %Cardiac Ultrasound
\titlerunning{Neural Multi-Scale Self-Supervised Registration} % Multi-Scale 
% \author{Anonymous Authors}
% \institute{Paper ID: 220}
\author{Wentao Zhu\inst{1} \and Yufang Huang\inst{2} \and Mani A Vannan\inst{3} \and Shizhen Liu\inst{3} \and Daguang Xu\inst{1} \and Wei Fan\inst{4} \and Zhen Qian\inst{4} \and Xiaohui Xie\inst{5}}
\institute{NVIDIA \email{{wentaoz1}@ics.uci.edu} \and Lenovo Research \and Piedmont Heart Institute \and Tencent Medical AI Lab \and University of California, Irvine \email{xhx@ics.uci.edu}}
\authorrunning{W. Zhu et al.}
\maketitle              % typeset the header of the contribution
%myocardial; cardiomyopathy disease related with muscle; 
\begin{abstract}
Echocardiography has become routinely used in the diagnosis of cardiomyopathy and abnormal cardiac blood flow. However, manually measuring myocardial motion and cardiac blood flow from echocardiogram is time-consuming and error-prone. Computer algorithms that can automatically track and quantify myocardial motion and cardiac blood flow are highly sought after, but have not been very successful due to noise and high variability of echocardiography.  
In this work, we propose a neural multi-scale self-supervised registration (NMSR) method for automated myocardial and cardiac blood flow dense tracking. NMSR incorporates two novel components: 1) utilizing a deep neural net to parameterize the velocity field between two image frames, and 2) optimizing the parameters of the neural net in a sequential multi-scale fashion to account for large variations within the velocity field.  Experiments demonstrate that NMSR yields significantly better registration accuracy than the state-of-the-art methods, such as advanced normalization tools (ANTs) and VoxelMorph,  for both myocardial and cardiac blood flow dense tracking. Our approach promises to provide a fully automated method for fast and accurate analyses of echocardiograms. 
%based on the quantitative metrics and visualizations. %evaluation The NMSR can potentially be of great use to radiologists in their effort to reduce the time on reading ultrasound images.

% We employ neural self-supervised optimization to acquire accurate velocity estimation that reduces the gap between the velocity prediction on training set and that on test set. The multi-scale registration framework facilitates the cardiac blood flow dense tracking of large variations on velocities because it alleviates the over-optimization of similarity function. 

% \vspace{-0.3cm}
 % \and Cardiac ultrasound screening
\end{abstract}
%
%
%
% \vspace{-1.2cm}
\section{Introduction}\label{sec:intro}
% \vspace{-0.4cm}
Cardiovascular diseases are the leading cause of mortality and morbidity globally. Echocardiogram is a non-invasive ultrasonic imaging test that exams the heart in motion. Because of the relatively low cost and ease of access, it has become a routinely used diagnostic tool in cardiology. Echocardiogram provides a wealth of information. Besides the assessment of the geometry and pumping capacity of the heart, echocardiogram has been used for the evaluation of the global and regional myocardial contractibility using speckle tracking techniques \cite{carasso2012velocity}. %, such as velocity vector imaging However, due to obesity and lung comorbidities, the use of contrast agents in echocardiogram has been proposed.

The use of contrast agents in echocardiogram has been proposed to further improve the imaging quality of echocardiogram. The advantages of using contrast include 1) a sharper demarcation between the left ventricular cavity and the myocardium, and 2) easier detection of sludges and thrombi in the heart. In addition, a high-power ultrasound beam produces a swirling pattern in the contrast-enhanced blood pool, which facilitates the visualization of the ventricular flow vortex. Unlike the conventional color Doppler technique, which estimates the flow velocity along a specific angle, the vortex imaging in contrast echocardiogram enables flow visualization and quantification in 2D, which is an important hemodynamic indicator of the heart's pumping performance and efficiency \cite{abe2013contrast}. In this work, we aim to achieve dense tracking of the myocardium in non-contrast echocardiogram and the blood flow in contrast echocardiogram, in an effort to make echocardiogram a one-stop shop for both myocardial contractibility assessment and ventricular flow pattern analysis.%\cite{senior2009contrast}

% From world health organization (WHO), cardiovascular diseases (CVDs) are the number one cause of death globally and more people die annually from CVDs than from other causes. CVDs cause 31\% of all global deaths which are about 17.9 million people died in 2016. And 85\% of deaths caused by CVDs are due to heart attack and stroke. Due to low cost and real time acquisition, echocardiography has become a widely used tool to find out if there is an abnormality of the heart that could lead to stroke and other cardiac diseases. 

% The study of cardiac motion can assist the diagnosis of several cardiac pathologies such as myocardial quantification and diastolic dysfunction \cite{geyer2010assessment}. Although Doppler technique can provide velocity estimation along a specific angle, it is limited by inaccuracy due to aliasing and frequency-dependent attenuation which prevent Doppler technique from becoming a standard daily practice. On the other hand, the speckle tracking can estimate velocity in the whole image without aliasing and angle dependent issues \cite{trahey1987angle}. In this work, we focus on myocardial and cardiac blood dense tracking on echocardiograms.
There are many previous methods on deformable registration that can be utilized for dense tracking in echocardiograms \cite{curiale2016influence}. Traditional registration methods are based on the optimization in registration field space such as elastic-type models \cite{bajcsy1989multiresolution,shen2002hammer}, FFD \cite{rueckert1999nonrigid}, Demons \cite{thirion1998image} and statistical parametric mapping \cite{ashburner2000voxel}. Diffeomorphic transformations preserve topology and many methods are derived from them such as LDDMM \cite{beg2005computing} and SyN \cite{avants2008symmetric}. The optimization of these traditional methods typically require substantial time. Deep learning based registration methods usually rely on ground truth of registration field \cite{rohe2017svf,sokooti2017nonrigid}. Recent unsupervised deep learning based registrations, such as VoxelMorph \cite{balakrishnan2018unsupervised}, are facilitated by the spatial transformer network \cite{de2017end,jaderberg2015spatial}, and the VoxelMorph is further extended to diffeomorphic transformation and Bayesian framework \cite{dalca2018unsupervised}. Adversarial similarity network adds an extra discriminator and uses adversarial training to improve the unsupervised registration \cite{fan2018adversarial}. These purely learning based methods cannot be directly applied to ultrasound images for velocity estimation especially for vortex detection in cardiac blood flow because of great noise in echocardiogram, large velocity variations of cardiac blood flow and large amounts of missing and new blood within the ultrasound plane. 

Inspired by the improvement of multi-scale registration and neural network parameterized  optimization \cite{curiale2016influence}, we propose a neural self-supervised optimization based multi-scale framework for dense tracking in echocardiograms as illustrated in Fig. \ref{fig:framework}. The neural self-supervised optimization yields accurate velocity field estimation by eliminating the gap between the estimation on the training set and that on the test set and alleviating the optimization difficulties such as local minima. Multi-scale strategy provides a sequential optimization pathway to the flow tracking that naturally emulates the formation of the fractal patterns of turbulent flow. %Multi-scale registration reduces the over-optimization on similarity function. li2017non , DARTEL \cite{ashburner2007fast},
% \vspace{-0.6cm}
\section{Neural Multi-Scale Self-Supervised Registration}\label{sec:nmsr}
\begin{figure}[t]
\begin{center}
    \includegraphics[width=\textwidth]{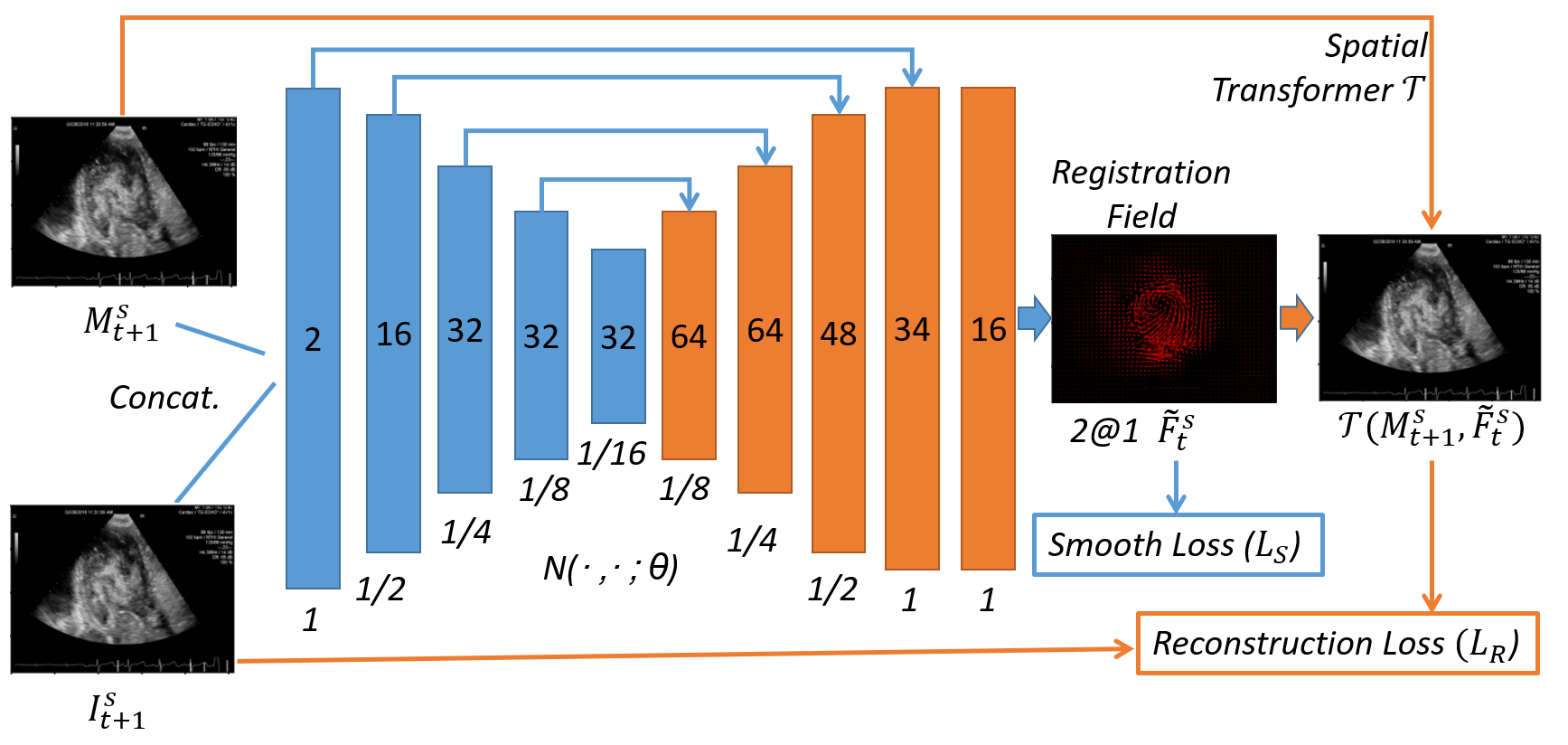}
\end{center}
% \vspace{-0.8cm}
\caption{\small Framework of the proposed neural multi-scale self-supervised registration. We employ U-Net to parameterize the registration function $\Tilde{\bm{F}}_t^s$ for velocity estimation between the reconstructed frame $\bm{M}_{t+1}^s$ from last scale $s/2$ and $t+1_{th}$ frame $\bm{I}_{t+1}^s$ in the scale $s$. Neural self-supervised optimization with reconstruction loss and smoothness loss is conducted to yield robust and accurate velocity field for each scale.}\label{fig:framework}%$\mathcal{L}_R$  $\mathcal{L}_S$ for each echocardiogram
% \vspace{-0.6cm}
\end{figure}

% \vspace{-0.4cm}
We denote an ultrasound sequence by $\bm{I} = \{\bm{I}_1, \bm{I}_2, \cdots, \bm{I}_t, \cdots, \bm{I}_n\}$, where $\bm{I}_t \in \mathbb{R}^{h \times w \times c}$ is the $t_{th}$ frame of $\bm{I}$, and $h$, $w$ and $c$ are image height, width and the number of channels respectively. In this work, we focus on calculating the registration field $\bm{F}_t$ as velocity field estimation which can be used by dense tracking for the two neighbor frames $\bm{I}_t$ and $\bm{I}_{t+1}$. We employ U-Net with skip connections to obtain the velocity field $\bm{F}_t$ \cite{ronneberger2015u}. The framework of neural multi-scale self-supervised registration (NMSR) is illustrated in Fig. \ref{fig:framework}.

Because the velocity of cardiac blood varies greatly, we extend the neural self-supervised registration to a multi-scale framework. We construct multi-scale U-Nets with scales $\{S, 2S, \cdots, s, \cdots, 1/2, 1\}$ to parameterize the registration fields of these scales where $S$ is the coarsest scale. In the registration of each scale $s$, we try to obtain the registration field $\Tilde{\bm{F}}_t^s$ by which we transform the reconstructed frame $\mathcal{T}(\bm{I}_t, \bm{F}_t^{s/2})$ from the last scale $s/2$ to the $t+1_{th}$ frame $\bm{I}_{t+1}$, where $t=1,\cdots,n-1$ and $\mathcal{T}$ is a spatial transformer network \cite{jaderberg2015spatial}. If $s$ is the coarsest scale $S$, $\mathcal{T}(\bm{I}_t, \bm{F}_t^{s/2}) = \bm{I}_t$. Given an ultrasound sequence $\bm{I}$, we firstly resize the moving image $\mathcal{T}(\bm{I}_t, \bm{F}_t^{s/2})$ and the fixed image $\bm{I}_{t+1}$ to the current scale $s$ for data preparation as%to obtain the registration field% of scale $s$ by three steps, data preparation, neural self-supervised optimization and field fusion%$\bm{F}_t^{s/2}$ is an identity field and 
% \vspace{-0.2cm}
\begin{equation}\label{eq:nmsr}
\begin{aligned}
    \bm{M}_{t+1}^s = \mathcal{P}_1 ( \mathcal{T}({\bm{I}}_t, {\bm{F}}_t^{s/2}), s ), \quad \bm{I}_{t+1}^s = \mathcal{P}_1 (\bm{I}_{t+1}, s), 
\end{aligned}
% \vspace{-0.2cm}
\end{equation}
where $\bm{M}_{t+1}^s$ is the scaled reconstructed frame and $\mathcal{P}_1$ is the down-sampling.

In addition to the high variations of velocities for cardiac blood flow, the signal-to-noise ratio of ultrasound image is typically low which easily leads to inaccurate prediction of the registration field. This aggregates the difficulty of learning and inference in the multi-scale registration model because the error can be accumulated over scales and the noisy error along the high dimensional output space makes the learning even harder than learning in the ultrasound space which is empirically observed in our experiments. We propose neural self-supervised optimization based on one ultrasound sequence, considering that the optimization based methods can eliminate the undesirable generalization ability of purely learning based model, and the settings and motions are consistent within one ultrasound sequence. We use stochastic optimization for the neural parameterized registration to alleviate the optimization difficulties such as local minima in traditional registration methods \cite{kingma2014adam}. The neural self-supervised optimization for the current scale $s$ and ultrasound sequence $\bm{I}$ can be formulated
% \vspace{-0.2cm}
\begin{equation}
    \begin{aligned}
        &\Tilde{\bm{F}}_t^{s} = \bm{N}(\bm{M}_{t+1}^s, \bm{I}_{t+1}^s; \bm{\theta}^s),\\
    &\bm{\theta}^s_{\star} = \mathop{\arg \min}_{\bm{\theta}^s} \sum_{t=1}^{n-1} {\mathcal{L}_{R}( \mathcal{T }(\bm{M}_{t+1}^s, \Tilde{\bm{F}}_t^{s}), \bm{I}_{t+1}^s; \bm{\theta}^s) + \lambda \mathcal{L}_{S}(\Tilde{\bm{F}}_t^{s}; \bm{\theta}^s)}, \\
    \end{aligned}
    % \vspace{-0.2cm}
\end{equation}
where $\bm{N}$ is the neural network with parameters $\bm{\theta}^s$, $\mathcal{L}_{R}$ is the reconstruction loss to measure the similarity between the moving image and fixed image, $\mathcal{L}_{S}$ is the smoothness loss for the registration field, $\lambda$ is the trade-off between reconstruction loss and smoothness loss.

We use negative local cross-correlation loss as reconstruction loss and $L_2$ norm of registration field gradient as the smoothness loss. For clarity, we omit $s$ in the reconstruction loss $\mathcal{L}_{R}$ and write the two losses as%\textcolor{red}{The cross-correlation formula is incorrect.}
% \vspace{-0.2cm}
\begin{equation}
\begin{aligned}
    \mathcal{L}_{R}({\bm{M}}_{t}, \bm{I}_{t}; \bm{\theta}^s) &= -\sum_{\bm{p}} \frac{ \big( \sum_{\bm{p}_i} {(\bm{M}_{t}(\bm{p}_i) - \bm{M}_{t}^{\prime}(\bm{p}) ) (\bm{I}_{t}(\bm{p}_i) - \bm{I}_{t}^{\prime}(\bm{p}) )} \big)^2}{\sum_{\bm{p}_i} (\bm{M}_{t}(\bm{p}_i) - \bm{M}_{t}^{\prime}(\bm{p}) )^2   \sum_{\bm{p}_i} (\bm{I}_{t}(\bm{p}_i) - \bm{I}_{t}^{\prime}(\bm{p}) )^2 }, \\ 
    \mathcal{L}_{S}(\Tilde{\bm{F}}_t^{s}; \bm{\theta}^s) &= \| \nabla_x \Tilde{\bm{F}}_t^{s} \|_{L_2} + \| \nabla_y \Tilde{\bm{F}}_t^{s} \|_{L_2}, 
\end{aligned} % \mathcal{L}_{R}({\bm{M}}_{t}, \bm{I}_{t}; \bm{\theta}^s) &= -\sum_{\bm{p}} \frac{ \sum_{\bm{p}_i} {(\bm{M}_{t}(\bm{p}_i) - \bm{M}_{t}^{\prime}(\bm{p}) ) (\bm{I}_{t}(\bm{p}_i) - \bm{I}_{t}^{\prime}(\bm{p}) )} }{\sqrt{ \sum_{\bm{p}_i} (\bm{M}_{t}(\bm{p}_i) - \bm{M}_{t}^{\prime}(\bm{p}) )^2   \sum_{\bm{p}_i} (\bm{I}_{t}(\bm{p}_i) - \bm{I}_{t}^{\prime}(\bm{p}) )^2 }}, \\ 
% \vspace{-0.2cm}
\end{equation}
where $\bm{p}$ is the pixel position in the frame and $\bm{p}_i$ is the pixel position within a square with the center as $\bm{p}$, $\bm{M}_{t}^{\prime}(\bm{p})$ and $\bm{I}_{t}^{\prime}(\bm{p})$ are local means of pixel position $\bm{p}_i$ in $\bm{M}_{t}$ and $\bm{I}_{t}$ respectively. %The radius of the local mean calculation is set as 6 in the experiments. ${\hat{\bm{I}}}_{t+1} = \Tilde{\bm{F}}_t^{s}(\bm{M}_{t+1}^s)$

After the neural self-supervised optimization, we obtain the optimal parameters $\bm{\theta}^s_{\star}$ of the neural network and can calculate the velocity field $\Tilde{\bm{F}}_t^{s}$ for the reconstructed frame $\mathcal{T}({\bm{I}}_t, {\bm{F}}_t^{s/2})$ from the last scale $s/2$. We calculate the registration field $\bm{F}_t^{s}$ for the ultrasound frame $\bm{I}_t$ by combining registration field $\bm{F}_t^{s/2}$ of the last scale and intermediate field $\Tilde{\bm{F}}_t^{s}$. % as
% \vspace{-0.2cm}
% \begin{equation}
%     \bm{F}_t^{s} = \mathcal{P}_2 (\frac{1}{s} \Tilde{\bm{F}}_t^{s}, \frac{1}{s}) \circ \bm{F}_t^{s/2} = \mathcal{P}_2 (\frac{1}{s} \bm{N}(\bm{M}_{t+1}^s, \bm{I}_{t+1}^s; \bm{\theta}^s_{\star}), \frac{1}{s}) \circ \bm{F}_t^{s/2},
%     \vspace{-0.2cm}
% \end{equation}
% where $\mathcal{P}_2$ is the linear interpolation, and $\circ$ is to combine two registration fields. %We use spatial transformer network transforming $\bm{M}_{t+1}^s$ to reconstruct $\bm{I}_{t+1}^s$ through field $\Tilde{\bm{F}_t^{s}}$ \cite{jaderberg2015spatial}. 
For each pixel position $\bm{p}$ in the ultrasound frame $\bm{I}_t$, we can obtain the final position $\hat{\bm{p}}$ and the combined registration filed $\bm{F}_t^{s}$ by
% \vspace{-0.2cm}
\begin{equation}
 %\quad \bm{F}_t^{s} = \bm{F}_t^{s} \circ \bm{F}_t^{s/2}
    \bm{F}_t^{s}(\bm{p}) = \hat{\bm{p}} - \bm{p} = \bm{F}_t^{s/2}(\bm{p}) + \mathcal{P}_2 (\frac{1}{s} \Tilde{\bm{F}}_t^{s}, \frac{1}{s})(\bm{p} + \bm{F}_t^{s/2}(\bm{p}) ),
    % \vspace{-0.2cm}
\end{equation}
where $\mathcal{P}_2$ is the linear interpolation and we use linear intepolation to calculate the field for $\bm{p} + \bm{F}_t^{s/2}(\bm{p})$.
% We do not need to calculate the combined registration for the coarsest scale $S$. 
% \vspace{-0.5cm}
\section{Experiments}\label{sec:expe}
% \vspace{-0.3cm}
We collected echocardiograms from 19 patients with 3,052 frames in total for myocardial tracking, and contrast echocardiograms from 71 patients with 11,462 frames in total for cardiac blood tracking. For testing, we randomly choose three patients' echocardiograms with 291 frames for myocardial tracking, and three patients' echocardiograms with 216 frames for cardiac blood tracking from the two datasets. The rest of echocardiograms are used as training. All echocardiograms have no registration field ground truth. %to construct myocardial tracking and cardiac blood tracking datasets respectively

%Because the second and third channels consist of low intensity signals in each frame,
We only use the first channel of echocardiography images (e.g., treated as gray-scale images) with the pixel values normalized to be in $[0, 1]$. 
%by dividing by $255$. To remove the background and improve the robustness of registration, we conduct registration on region of pixel value within the range $[0.005, 0.995]$.  To obtain a smooth boundary, we further employ morphological dilation with a disk-shaped structure of radius of $16$ pixels. 
For cardiac blood tracking, we extract cardiac blood region by 1) creating masks of the left ventricular blood pool at the end of the systole and the end of the diastole, 2) using active contour model to fit 100 uniformly sampled spline points along a circle into the boundary of cardiac blood mask \cite{kass1988snakes}, 3) using linear interpolation to get 100 interpolated spline points for each frame, 4) using radial basis function in interpolation to get the final smooth cardiac blood boundary from the 100 spline points. Removing myocardial region is crucial to cardiac blood tracking. % extract myocardial region  creating cardiac blood masks for frame with contraction of the atria and frame with ventricular relaxation, which are the start frame and the end frame for one cardiac cycle

For unsupervised learning or self-supervised optimization, one of the main challenges is the model evaluation. Manually labeling the corresponding points for evaluation is time-consuming, laborious and inaccurate, because the total size of one frame is $1024 \times 768$ and the signal-to-noise ratio is low. Instead of using pixel position based evaluation metric, we use reconstruction based metrics, i.e., the mean square error (MSE) and the mean local cross correlation (Mean CC) with radius as 10. For the metrics in Table \ref{tab:comp}, we calculate the average MSE and Mean CC over all frame pairs $\bm{I}_t$ and $\bm{I}_{t+1}$ with the pixel value of range $[0, 1]$. For MSE and Mean CC of one frame pair, we take the average of square error and local cross correlation over the masked region.

\begin{table}[t]
\caption{\small Comparisons on myocardial and cardiac blood dense tracking among ANTs, VoxelMorph and NMSR. NMSR obtains the best performance.}\label{tab:comp}% on both myocardial and cardiac blood dense tracking. \textcolor{red}{The ANTs results are bad, even worse than VoxelMorph. Need explanation? }
% \vspace{-0.3cm}
\begin{tabular}{|c|c|c|c|c|}
\hline
\multirow{2}{*}{Methods} & \multicolumn{2}{c|}{Myocardial tracking} & \multicolumn{2}{c|}{Cardiac blood tracking} \\ \cline{2-5} 
                         & MSE (10\textsuperscript{-3})      & Mean CC (10\textsuperscript{-1})       & MSE (10\textsuperscript{-3})      & Mean CC (10\textsuperscript{-1}) \\ \hline
ANTs & 15.5179\textpm9.4637 & 3.1597\textpm1.3913 & 3.9344\textpm1.4660 & 4.2062\textpm1.0576 \\ \hline % no mask 8.4777\textpm4.7977 & 1.1237\textpm0.5220 & 11.1265\textpm2.9642
VoxelMorph & 1.2266\textpm0.5457 & 4.7363\textpm0.5457 & 5.8654\textpm1.6943 & 3.3462\textpm0.5891 \\ \hline \hline %1.7207\textpm0.7996 & 6.3032\textpm0.6703 1.3014\textpm0.3811 & 1.9989\textpm0.2730

NMSR & 1.3438\textpm0.6248 & 4.6270\textpm0.5340 & 5.9511\textpm1.6789 & 3.2228\textpm0.5754 \\ \hline %0.6858\textpm0.3179 & 2.2721\textpm0.3179
NMSR (1/8) & 1.7938\textpm0.5992 & 4.1428\textpm0.5066 & 6.8603\textpm1.4958 & 2.3314\textpm0.5497 \\ \hline %6.4645\textpm1.5685 & 2.6009\textpm0.4330%0.9605\textpm0.3053 & 2.0151\textpm0.2609
NMSR (1/4) & 1.8185\textpm0.5437 & 4.2746\textpm0.4700 & 5.6694\textpm1.3347 & 2.7566\textpm0.6186 \\ \hline %5.2521\textpm1.2351 & 2.8724\textpm0.4850%0.9711\textpm0.3390 & 2.0839\textpm0.2481
NMSR (1/2) & 1.5405\textpm0.4515 & 4.5508\textpm0.4601 & 4.6870\textpm1.0579 & 3.2854\textpm0.7191 \\ \hline %4.8306\textpm1.1222 & 3.1203\textpm0.5316%0.8243\textpm0.2818 & 2.1977\textpm0.2257
NMSR (1) & \bf{1.3204\textpm0.4058} & \bf{4.8954\textpm0.4173} & \bf{4.0337\textpm0.9903} & \bf{3.9781\textpm0.6917} \\ \hline \hline %\bf{4.4255\textpm1.0054} & \bf{3.3818\textpm0.5884}%0.7113\textpm0.2471 & 2.3833\textpm0.2124

% use vm as pretrain model
NMSR\textsubscript{V} & 1.2158\textpm0.5473 & 4.7809\textpm0.4952 & 5.7780\textpm1.6222 & 3.3942\textpm0.5868 \\ \hline %1.1989\textpm0.5457 & 4.7961\textpm0.4917 0.6747\textpm0.3106 & 2.2825\textpm0.2643
NMSR\textsubscript{V} (1/8) & 1.6937\textpm0.5616 & 4.2084\textpm0.5086 & 6.8108\textpm1.8189 & 2.3948\textpm0.5495 \\ \hline %6.7600\textpm1.7667 & 2.5501\textpm0.4150 %1.8905\textpm0.7281 & 4.0972\textpm0.4885
NMSR\textsubscript{V} (1/4) & 1.7022\textpm0.4434 & 4.3377\textpm0.4674 & 5.6577\textpm1.3317 & 2.7109\textpm0.5988 \\ \hline %5.3769\textpm1.2617 & 2.8546\textpm0.4770%1.8884\textpm0.6503 & 4.2882\textpm0.4900
NMSR\textsubscript{V} (1/2) & 1.4363\textpm0.3625 & 4.6459\textpm0.4451 & 4.6228\textpm1.1381 & 3.3155\textpm0.6691 \\ \hline %4.7271\textpm1.0669 & 3.1432\textpm0.5340%1.5871\textpm0.5322 & 4.6092\textpm0.4559
NMSR\textsubscript{V} (1) & 1.2420\textpm0.3041 & 5.0280\textpm0.3922 & 3.9287\textpm1.0209 & 4.1209\textpm0.6188 \\ \hline %4.4322\textpm1.0108 & 3.3715\textpm0.5717%1.3821\textpm0.4689 & 4.9839\textpm0.4094

% use vm and increamental learning 
% do not use mask to calculate score for muscle
%DSSR\textsubscript{VI} (1/8) & 2.0716\textpm0.5581 & 3.8942\textpm0.3536 & 6.8242\textpm1.8483 & 2.5267\textpm0.4275 \\ \hline
NMSR\textsubscript{VI} (1/4) & 1.3504\textpm0.4885 & 4.4140\textpm0.5005 & 5.6106\textpm1.2424 & 2.7331\textpm0.6779  \\ \hline %5.4153\textpm1.3011 & 2.8468\textpm0.4859 %1.7778\textpm0.4629 & 4.3044\textpm0.4593
NMSR\textsubscript{VI} (1/2) & 1.0929\textpm0.3775 & 4.7155\textpm0.4569 & 4.5089\textpm1.0272 & 3.3879\textpm0.7524 \\ \hline %4.7179\textpm1.1069 & 3.1406\textpm0.5361 %1.4927\textpm0.4022 & 4.6216\textpm0.4374
NMSR\textsubscript{VI} (1) & \bf{0.9206\textpm0.3236} & \bf{5.0881\textpm0.4107} & \bf{3.7944\textpm0.9384} & \bf{4.2519\textpm0.7181} \\  \hline %\bf{4.3626\textpm1.0224} & \bf{3.4456\textpm0.5857} %1.2825\textpm0.3367 & \bf{4.9986\textpm0.3927}
\end{tabular}
% \vspace{-0.9cm}
\end{table}

We compare our approach to symmetric normalization (SyN) with cross-correlation similarity as implemented in ANTs \cite{avants2008symmetric,avants2009advanced}, and VoxelMorph \cite{balakrishnan2018unsupervised}. ANTs (SyN) is a traditional optimization based method and VoxelMorph is a deep learning based unsupervised registration with a similar network structure and loss function as NMSR. For the purpose of ablation studies, we report the results of {\bf{1)}} NMSR, using only one scale and neural self-supervised optimization, {\bf{2)}} NMSR (1/8), the coarsest scale of neural multi-scale self-supervised registration, {\bf{3)}} NMSR (1/4), the second coarsest scale of NMSR, {\bf{4)}} NMSR (1/2), the third coarsest scale of NMSR, {\bf{5)}} NMSR (1), the final scale of NMSR after sequential multi-scale optimization, {\bf{6)}} NMSR\textsubscript{V}, NMSR\textsubscript{V} (*), which use training-data optimized NMSR as an initialization and conduct self-supervised neural optimization afterwards, and {\bf{7)}} NMSR\textsubscript{VI} (*), which use the neural weights from the last scale as an initialization.% and conduct self-supervised neural optimization afterwards.

For all these multi-scale based methods, we use four different scales $1/8$, $1/4$, $1/2$ and $1$. We use radius of 6 pixels for the local cross-correlation loss in all these methods. We set the number of optimization steps to 200 for each scale in ANTs. We use learning rate of $1\times10^{-3}$ and Adam optimizer to update the weights in neural networks for both NMSR and VoxelMorph \cite{kingma2014adam}. The $\lambda$ is set to be 10. We set the number of optimization steps to $3500$ per ultrasound sequence for the self-supervised neural optimization, and set the number of iterations to $3500 \times$ the number of training ultrasound sequences for VoxelMorph.
\begin{figure}[t]
	\begin{center}
		\begin{minipage}{0.325\linewidth}
			\includegraphics[width=\textwidth]{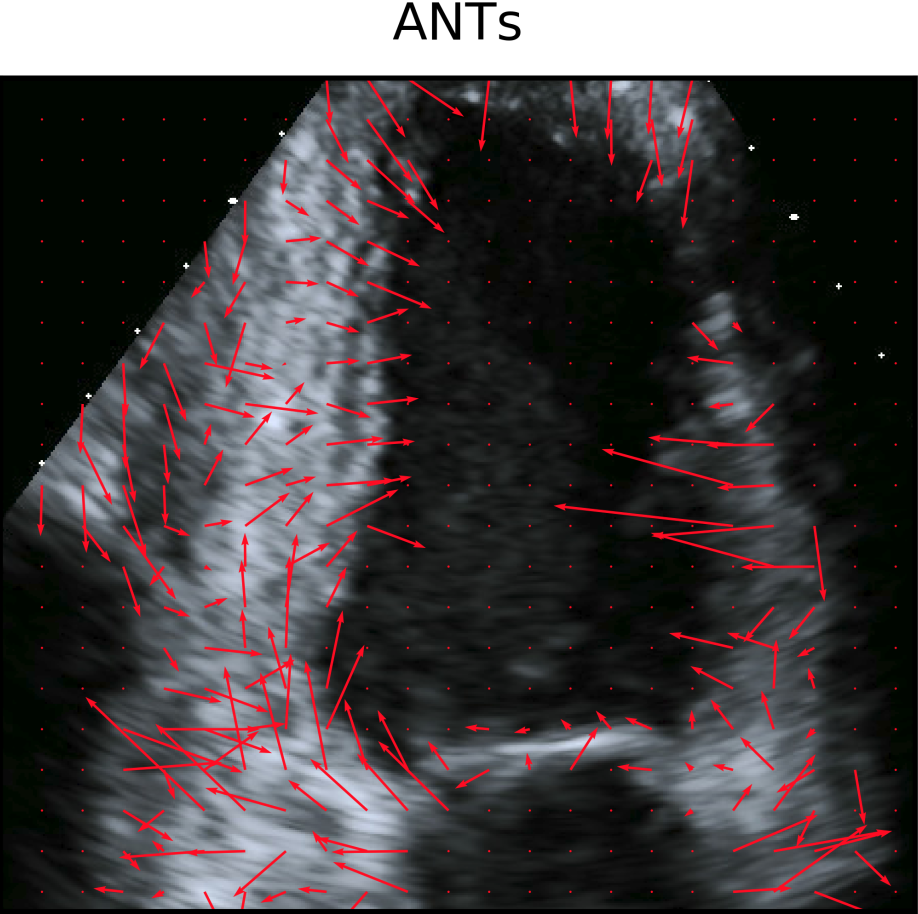}
		\end{minipage}
		\begin{minipage}{0.325\linewidth}
			\includegraphics[width=\textwidth]{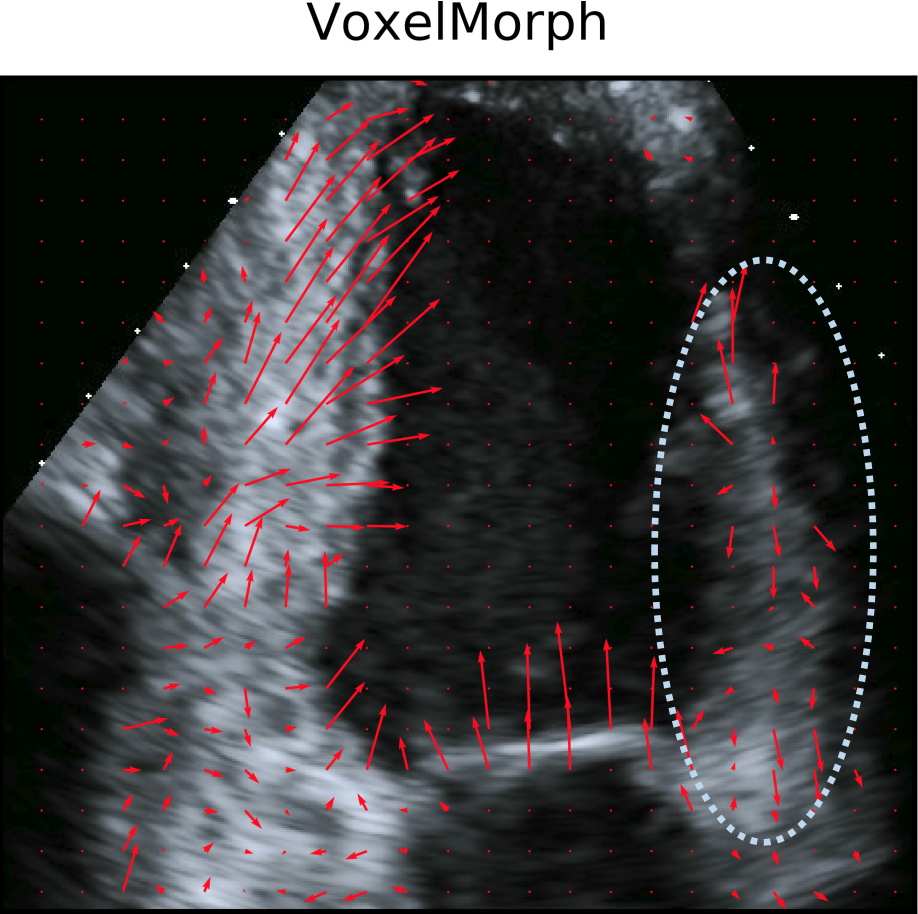}
		\end{minipage}
		\begin{minipage}{0.325\linewidth}
			\includegraphics[width=\textwidth]{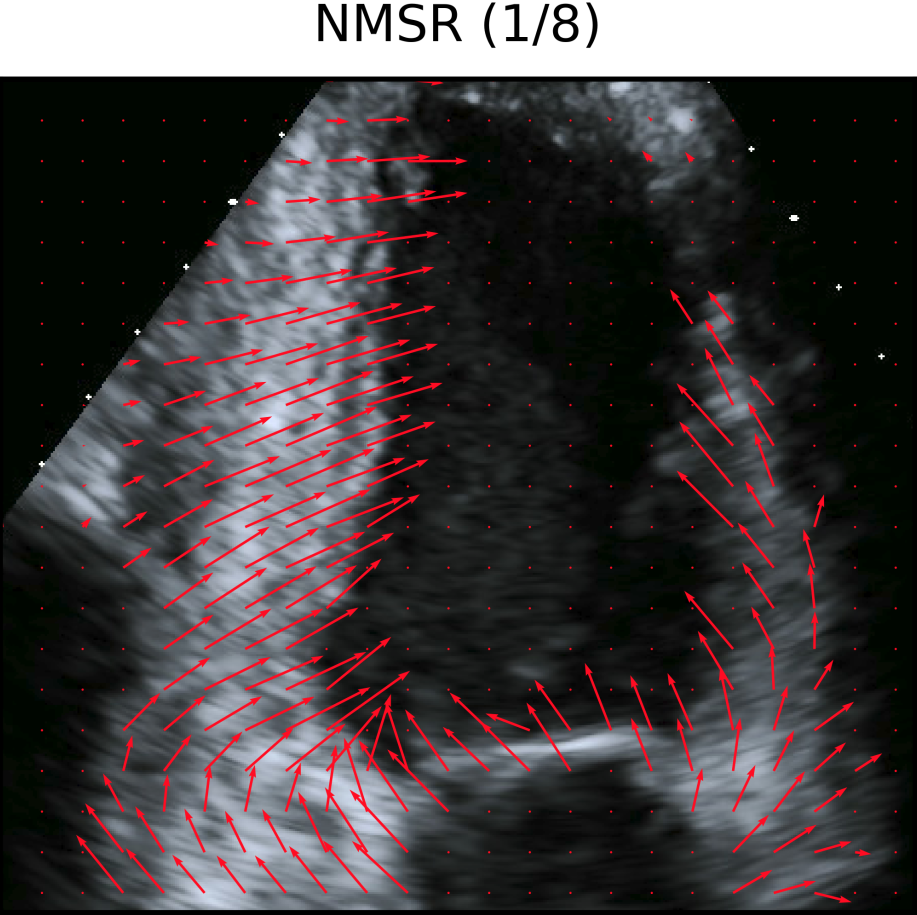}
		\end{minipage} \\
		
		\begin{minipage}{0.325\linewidth}
			\includegraphics[width=\textwidth]{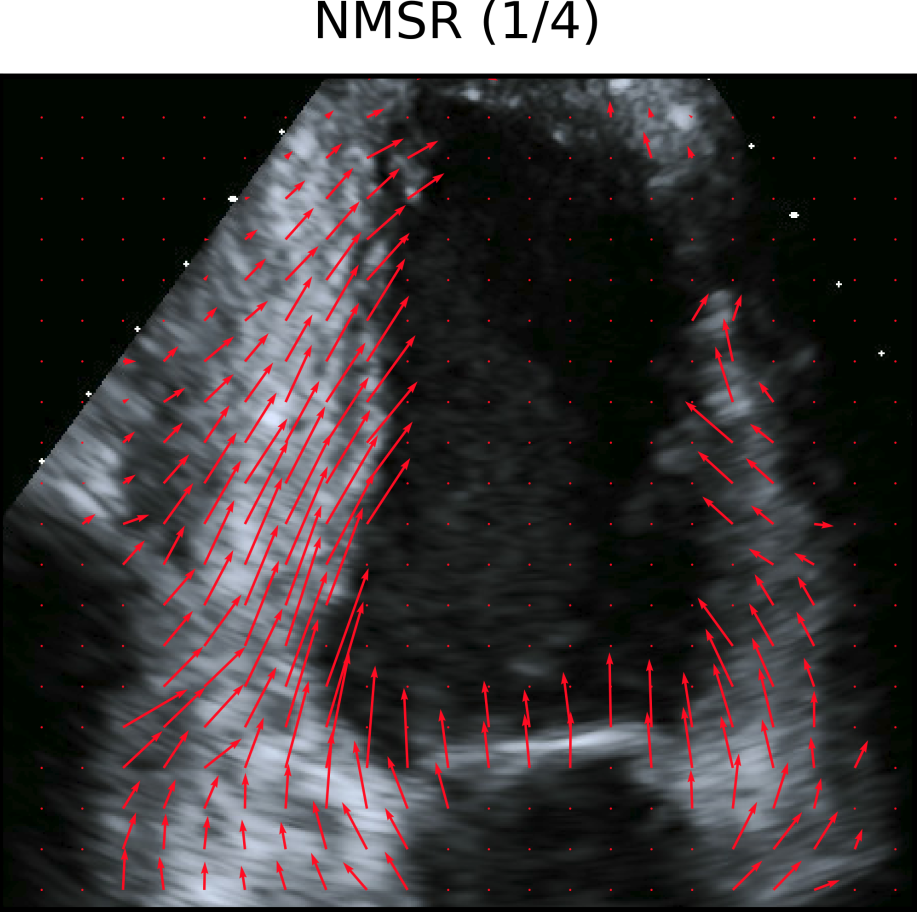}
		\end{minipage}
		\begin{minipage}{0.325\linewidth}
			\includegraphics[width=\textwidth]{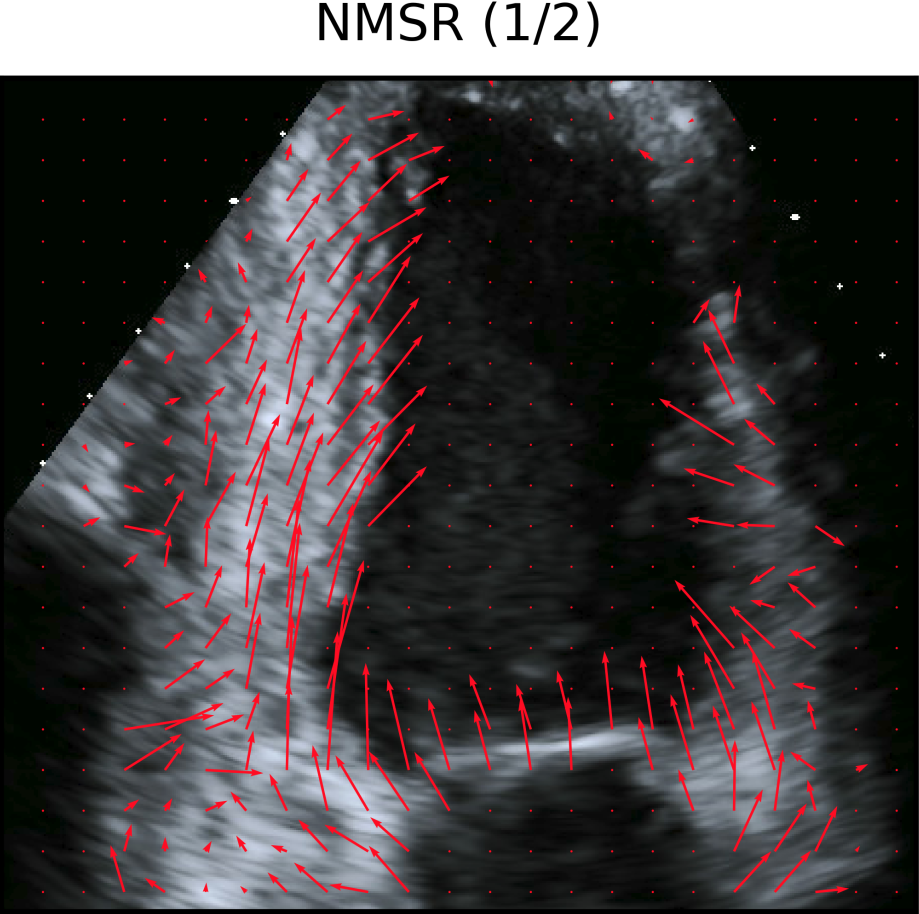}
		\end{minipage}
		\begin{minipage}{0.325\linewidth}
			\includegraphics[width=\textwidth]{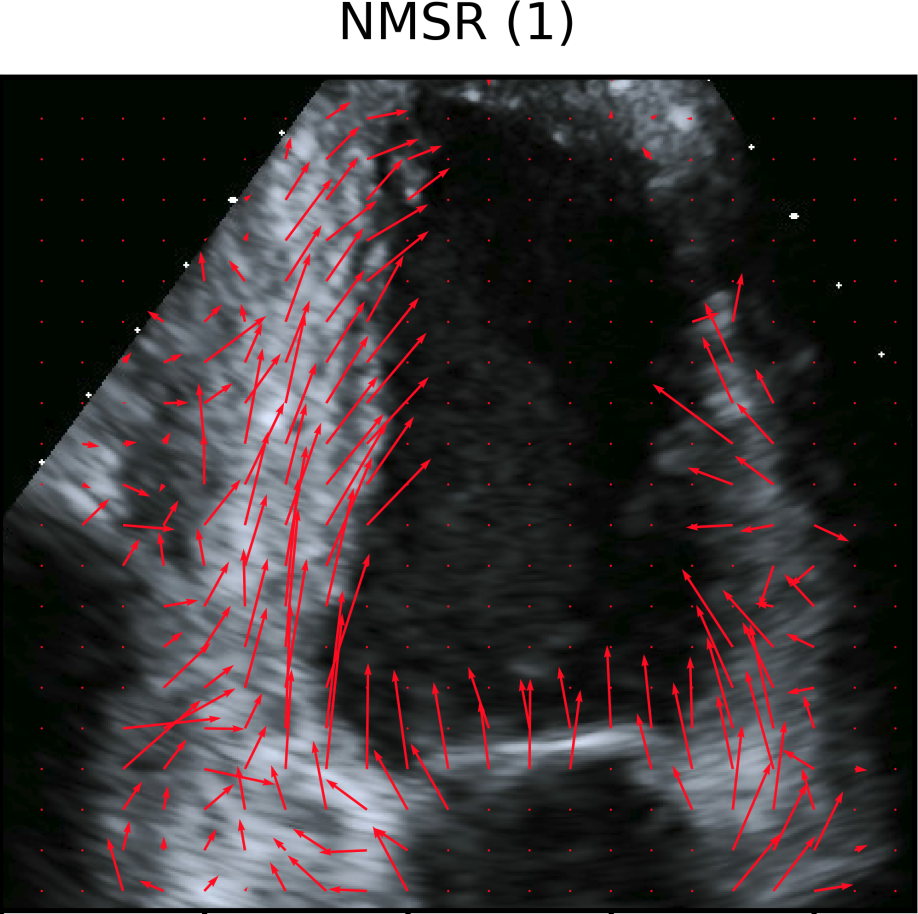}
		\end{minipage} 
	\end{center}
% 	\vspace{-0.5cm}
	\caption{\small Visualization of myocardial tracking based on ANTs, VoxelMorph and NMSR. The right side of velocity field by VoxelMorph is inaccurate. NMSR yields smooth, coarse-to-fine and accurate velocity estimation.}\label{fig:muscle}%field  ANTs produces chaotic registration field.
% 	\vspace{-0.7cm}
\end{figure}

\begin{figure}[t]
	\begin{center}
		\begin{minipage}{0.325\linewidth}
			\includegraphics[width=\textwidth]{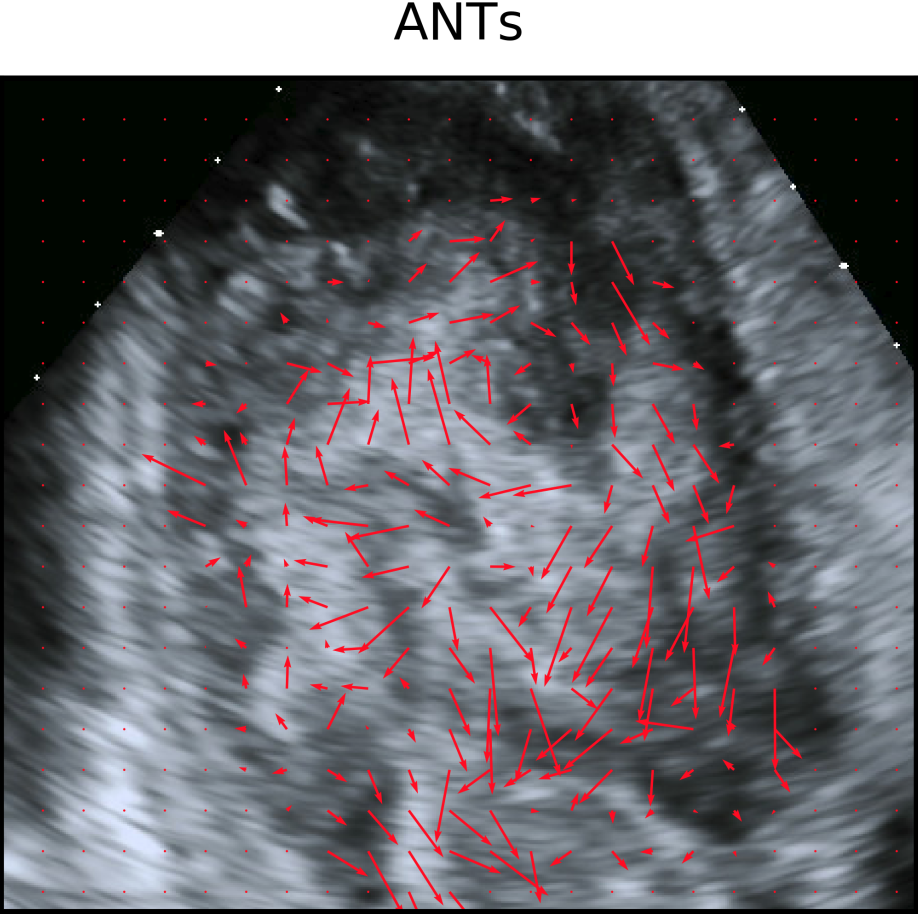}
		\end{minipage}
		\begin{minipage}{0.325\linewidth}
			\includegraphics[width=\textwidth]{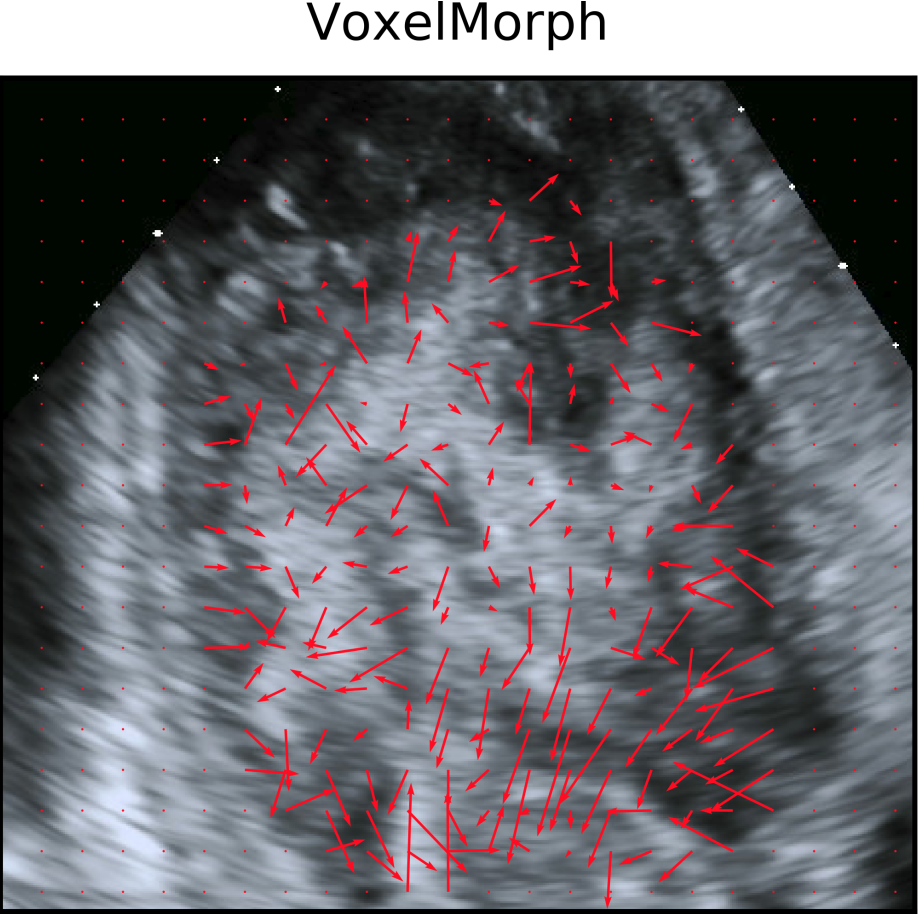}
		\end{minipage}
		\begin{minipage}{0.325\linewidth}
			\includegraphics[width=\textwidth]{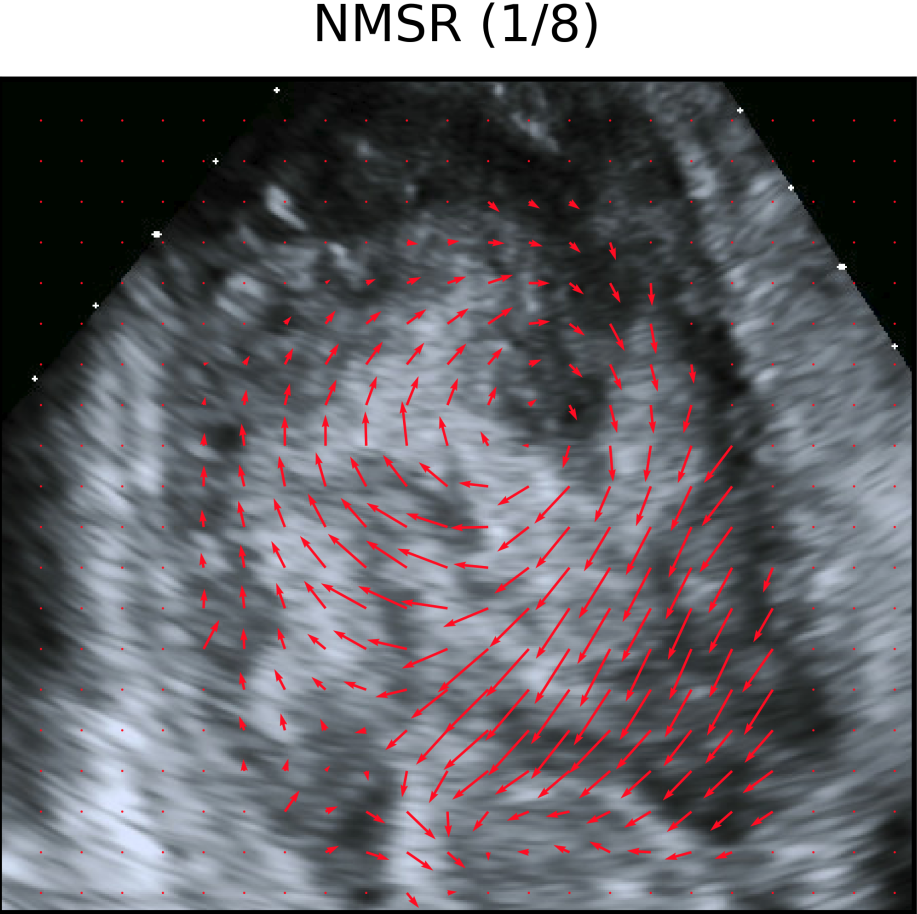}
		\end{minipage} \\
		
		\begin{minipage}{0.325\linewidth}
			\includegraphics[width=\textwidth]{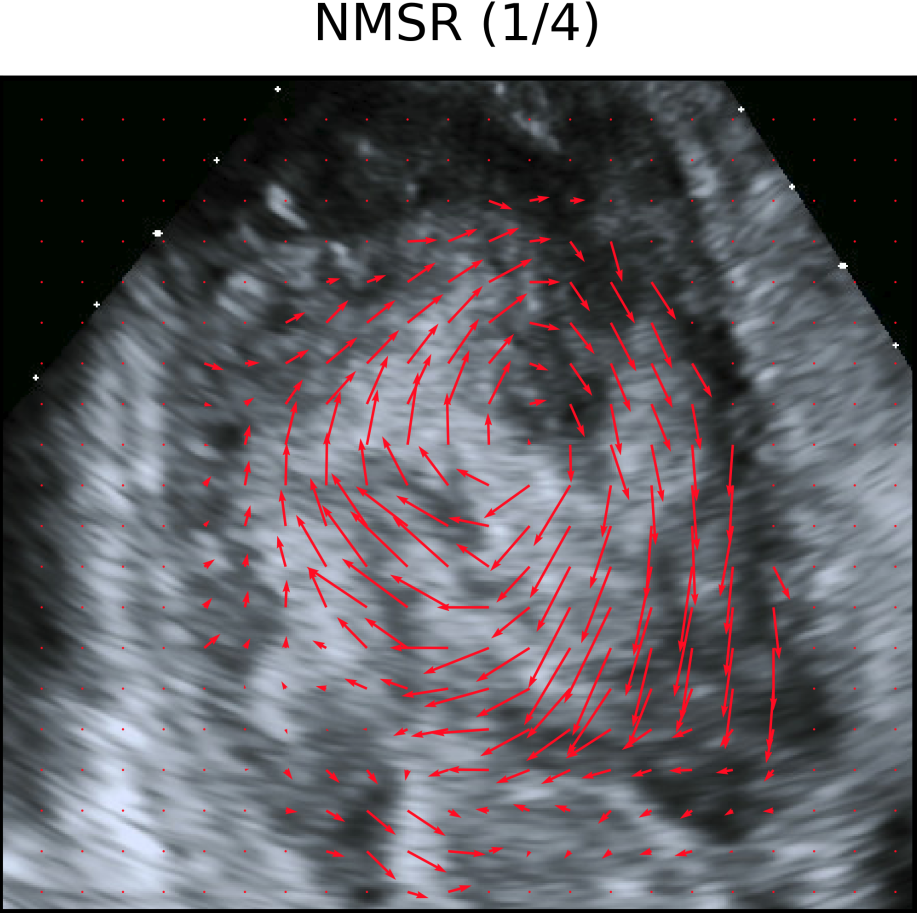}
		\end{minipage}
		\begin{minipage}{0.325\linewidth}
			\includegraphics[width=\textwidth]{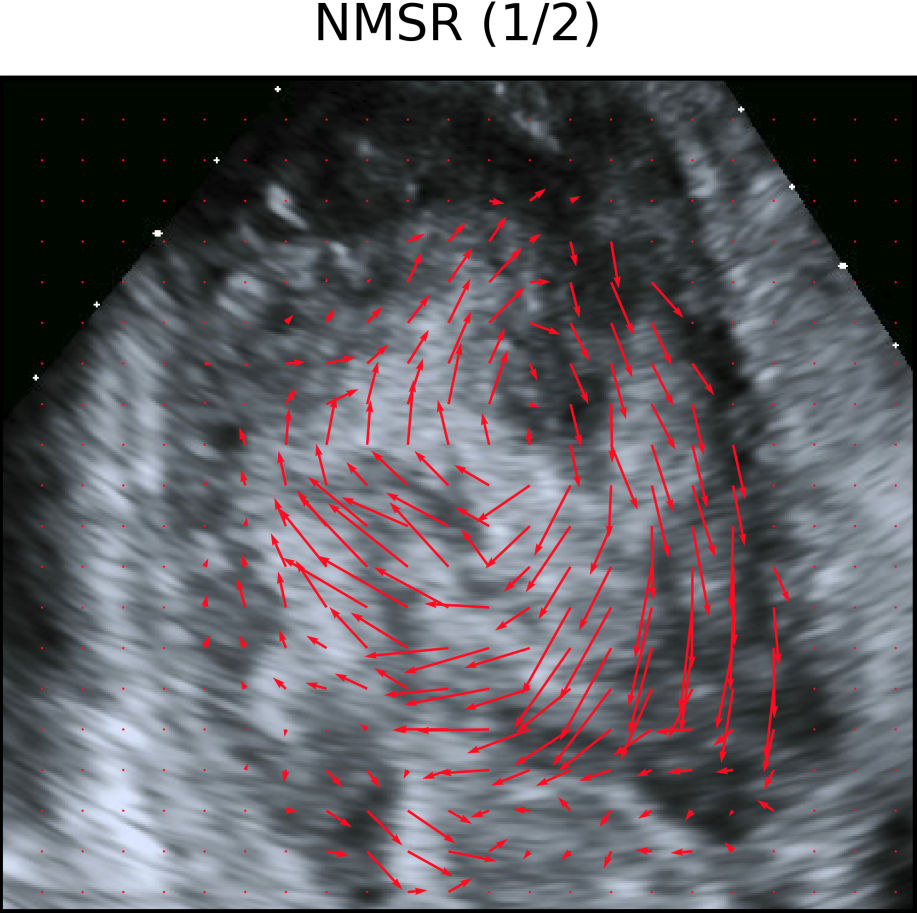}
		\end{minipage}
		\begin{minipage}{0.325\linewidth}
			\includegraphics[width=\textwidth]{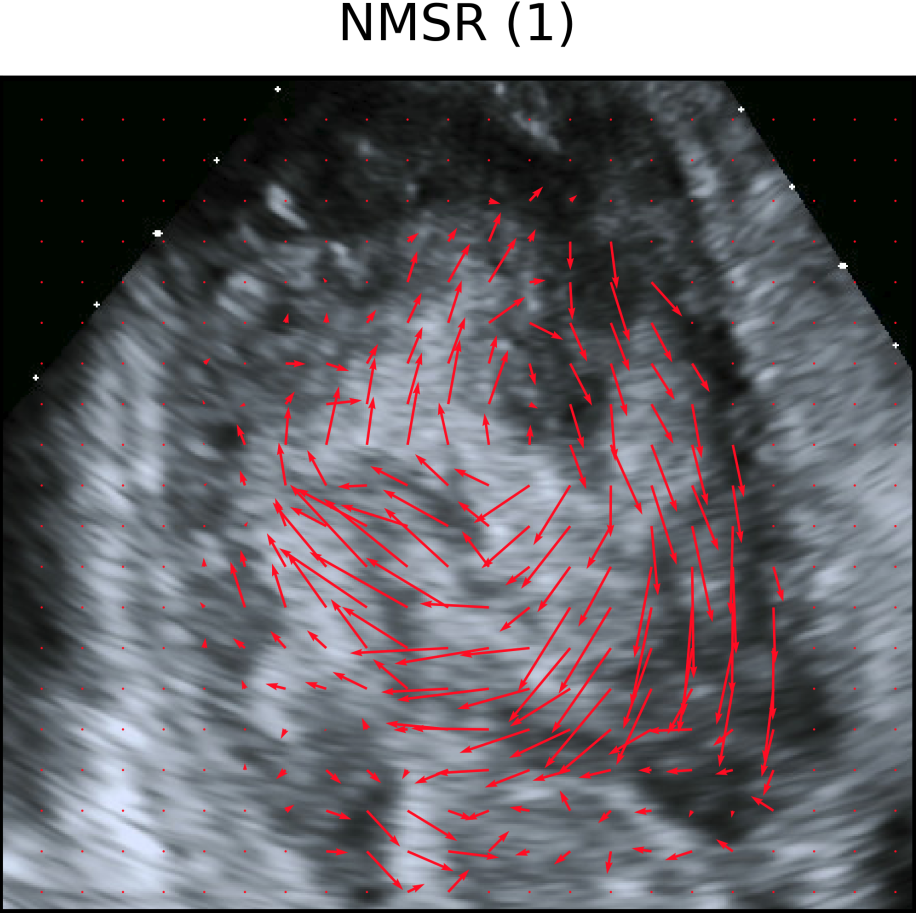}
		\end{minipage}
	\end{center}
% 	\vspace{-0.5cm}
	\caption{\small Visualization of cardiac blood tracking based on ANTs, VoxelMorph and NMSR. ANTs and VoxelMorph produce noisy velocity fields. NMSR successfully detects the vortex and yields coarse-to-fine and smooth velocity field.}\label{fig:blood}
% 	\vspace{-0.8cm}
\end{figure}

Quantitative comparison results of our model against the state-of-the-arts on both myocardial and cardiac blood flow dense tracking are shown in Table \ref{tab:comp}. We highlight the following a few observations:  1) NMSR\textsubscript{VI} (1) achieves the best performances, and outperforms ANTs in terms of both MSE and mean CC on both tasks, likely due to the representation and optimization efficiency of deep neural nets;  2) NMSR\textsubscript{V} yields consistently better results than VoxelMorph, demonstrating the efficacy of self-supervised optimization during the test phase for improving velocity field estimation and reducing the estimation gap between training and testing; 3) NMSR (1) achieves better performance than NMSR on all experiments, demonstrating the benefit of sequential multi-scale optimization in echocardiogram registration.  The multi-scale scheme alleviates the over-optimization of reconstruction loss, which can be visually noticed from Fig. \ref{fig:muscle} and \ref{fig:blood}; and 4) NMSR\textsubscript{V} (1) obtains better performance than NMSR (1), illustrating the benefit of using pretrained models as an initialization for NMSR, further confirmed by the better performance of NMSR\textsubscript{VI} over NMSR\textsubscript{V}. %From Table \ref{tab:comp}, for the two evaluation metrics on both myocardial and cardiac blood tracking  has a good capacity to model the registration field

% 96_94_144 7  96 94 144 9 
{\bf{Visualizations}} We visualize the myocardial tracking results based on ANTs, VoxelMorph and the intermediate registration fields from NMSR\textsubscript{VI} with four different scales in Fig. \ref{fig:muscle}. We randomly choose one frame from these ultrasound images. From Fig. \ref{fig:muscle}, we note that the registration field from ANTs is noisy, and the velocity direction from VoxelMorph for the right myocardial is incorrect. By contrast, NMSR (1/8) produces the smoothest registration field, and NMSR (1) generates more detailed velocity estimation that preserves both large and low-scale velocity variations.  The coarse-to-fine results illustrate that the multi-scale optimization scheme coupled with deep neural nets can be very effective in dealing with the highly challenging case of image registration in echocardiograms.

%also
We visualize the cardiac blood tracking results in Fig. \ref{fig:blood}. From the registration fields from ANTs and VoxelMorph, we cannot easily recognize the vortex in cardiac blood flow. By contrast, the vortex flow pattern from NMSR is readily recognizable. The general vortex pattern is apparent from the coarsest level registration by NMSR (1/8), followed by finer-scale registrations to introduce details of local velocity field variations. The final velocity field produced by NMSR (1) includes both easily recognizable vortex flow, as well as details of local field variations. 

%NMSR (1/8) produces the general velocity direction to ensure that the following registration field not fall too far from the coarsest registration field. The neural self-supervised multi-scale framework is noise tolerant and  works on the complicated blood flow prediction. %based on ANTs, VoxelMorph and the intermediate registration fields from NMSR with four different scales

{\bf{Computational Cost}} For ANTs, the average computational time is 214.10 \textpm54.04 seconds for the registration of two consecutive frames on 12 processors of Intel i7-6850K CPU @ 3.60GHz. For an ultrasound sequence of 50 frames, the computational time is about {\bf{three hours}} for ANTs. Because the inference of VoxelMorph only relies on one feed forward pass of deep neural network, the average computational time is 0.11\textpm0.47 seconds for one pair frames on one NVIDIA 1080 Ti GPU. The NMSR takes 279.97, 101.65, 68.79, 66.09 seconds for neural self-supervised optimization with the scale 1, 1/2, 1/4, 1/8 respectively on one ultrasound sequence of 49 frames by one NVIDIA 1080 Ti GPU. The NMSR takes less than {\bf{nine minutes}} on the neural self-supervised optimization in total for one ultrasound sequence, achieving {\bf{20}} times speedup over ANTs.% even using four scales.
% \vspace{-0.6cm}
\section{Conclusion}\label{sec:conc}
% \vspace{-0.3cm}
In this work, we propose a novel framework, neural multi-scale self-supervised registration (NMSR), for both myocardial and cardiac blood dense tracking. To produce accurate velocity estimation from noisy ultrasound images and reduce the estimation gap between training and testing, we incorporate self-supervised optimization in the registration framework. To handle large variations of velocity fields in echocardiogram tracking, a multi-scale scheme is integrated into the proposed framework to reduce the over-optimization of similarity functions. Our proposed method consistently outperforms state-of-the-art methods on both myocardial and cardiac blood flow dense tracking. With further improvements on model and optimization, to consider for example other loss functions and extend it to diffeomorphic registrations, it seems plausible to have a fully automated method for echocardiogram analysis. 

%In the future, we will collect echocardiogram with registration ground truth to further study the estimated position error of these methods. % We employ the advantage of neural optimization to alleviate the optimization difficulties such as avoiding local minima.
% \vspace{-0.5cm}
\bibliographystyle{splncs04}
% \vspace{-0.5cm}
 \small {\bibliography{mybibliography}}
\end{document}